\documentclass[10pt,twocolumn,letterpaper]{article}

\usepackage{wacv}
\usepackage{times}
\usepackage{epsfig}
\usepackage{graphicx}
\usepackage{amsmath}
\usepackage{amssymb}
\usepackage{mathtools, nccmath}
\usepackage{subcaption}

\wacvfinalcopy 

\def\wacvPaperID{1254} 

\ifwacvfinal\fi
\begin{document}
\DeclareGraphicsExtensions{.pdf}
\title{LR-to-HR Face Hallucination with an Adversarial Progressive Attribute-Induced Network}

\author{Nitin Balachandran \\
University of Maryland\\
{\tt\small nbalacha@umd.edu}
\and
Jun-Chen Cheng \\
Academia Sinica\\
\and
Rama Chellappa \\
Johns Hopkins University\\
}

\maketitle
\ifwacvfinal\thispagestyle{empty}\fi

\begin{abstract}

Face super-resolution is a challenging and highly ill-posed problem since a low-resolution (LR) face image may correspond to multiple high-resolution (HR) ones during the hallucination process and cause a dramatic identity change for the final super-resolved results. Thus, to address this problem, we propose an end-to-end progressive learning framework incorporating facial attributes and enforcing additional supervision from multi-scale discriminators. By incorporating facial attributes into the learning process and progressively resolving the facial image, the mapping between LR and HR images is constrained more, and this significantly helps to reduce the ambiguity and uncertainty in one-to-many mapping. In addition, we conduct thorough evaluations on the CelebA dataset following the settings of previous works (\emph{i.e.,} super-resolving by a factor of 8$\times$ from tiny 16$\times$16-pixel face images.), and the results demonstrate that the proposed approach can yield satisfactory face hallucination images outperforming other state-of-the-art approaches.

\end{abstract}

\section{Introduction}

Face analytics \cite{face_rec_survey} is one of the active research areas of computer vision  with plenty of applications, including face recognition for access authentication, fiducial point detection for virtual makeup, etc. Pose, poor illumination, obscured/occluded faces, and other conditions  usually make understanding the identity or the content of a face a challenging task. Among them, the resolution of a face is one of most difficult factors which significantly hinders the detection and recognition of the face image if it is too small. To address this challenge, mapping low resolution (LR) face images to high resolution (HR) face images, better known as face hallucination, has been gaining popularity. Current face hallucination approaches \cite{DBLP:journals/corr/ZhuLLT16,Yu2016UltraResolvingFI,Yu2017HallucinatingVL,Cao2017AttentionAwareFH} try to super resolve LR facial images into HR facial images by learning the mapping between these two spaces using relevant datasets.

\begin{figure}[t!]
    \centering
    \includegraphics[height=4cm]{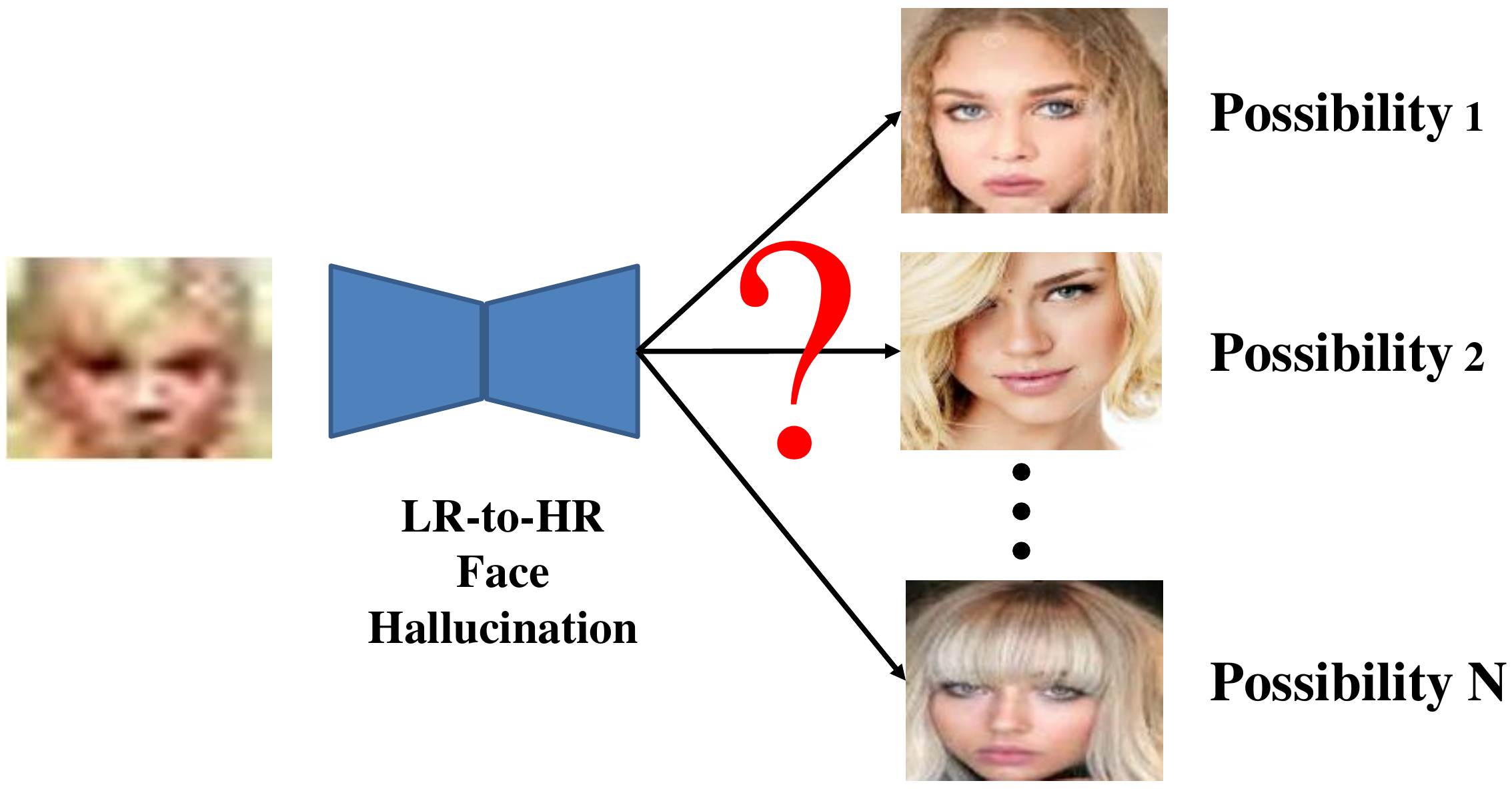}
    \caption{An illustrative example for the situation of hallucinating a HR image from its LR one, there exists an issue of one-to-many mapping due to the ill-posed nature of
    image super-resolution. It may lead to the problem of identity change during the process of LR-to-HR hallucination. 
    }
    \label{fig:ambiguity}
\end{figure}

\begin{figure*}[t]
    \centering
    \includegraphics[width=1\textwidth,height=11cm]{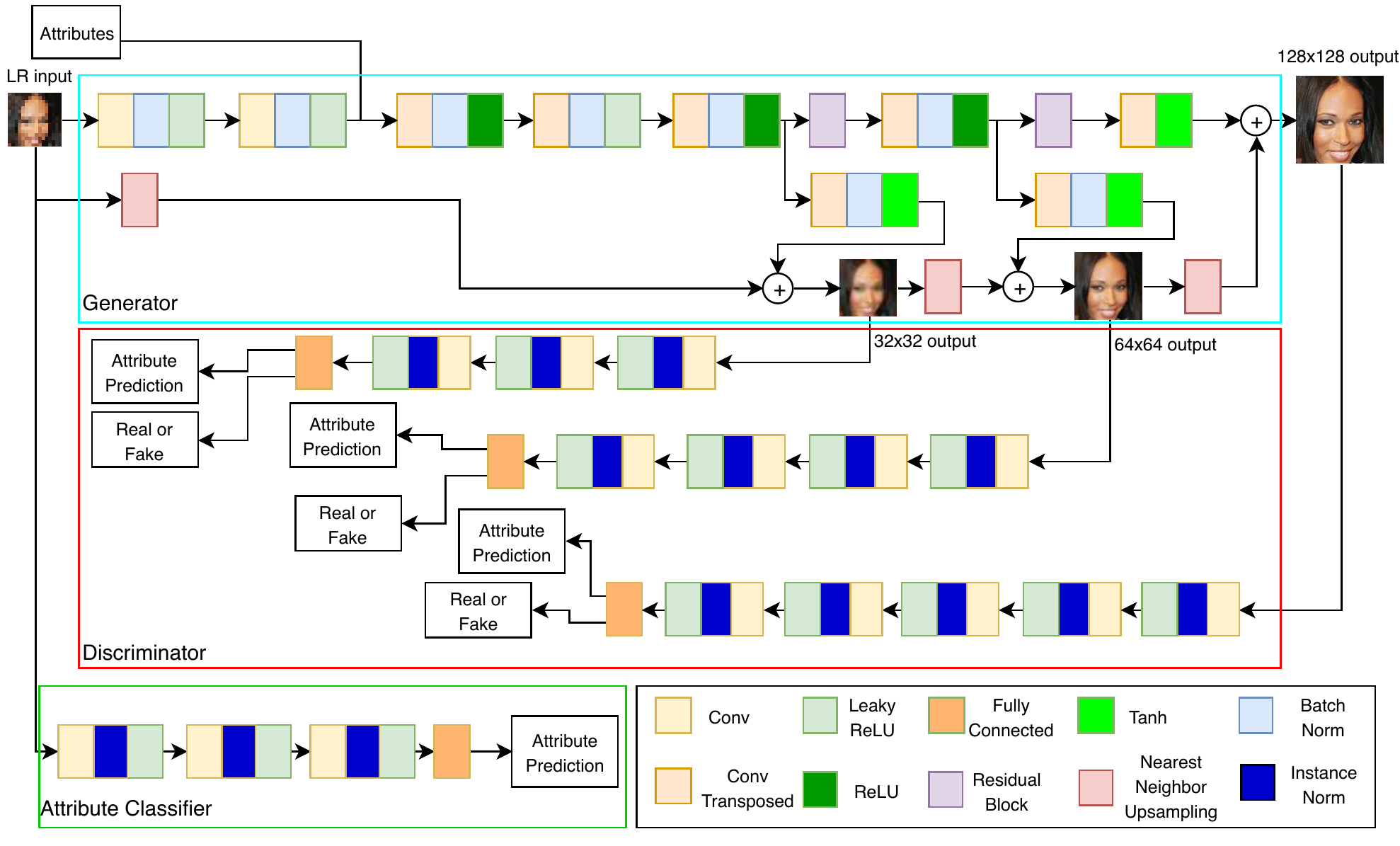}
    \caption{An overview of the architecture of our model. The model includes a generative network that takes in a LR face image and its corresponding attributes and produces a HR resolution face image of it. Since our model is progressively trained, the generative model produces an intermediary resolution face image at each stage. In addition our model has a discriminative network for each stage whose input is that stage's face image and predicts its corresponding attributes and if it is real or fake. Finally, there is an attribute classifier that takes in the LR face image and tries to predict the attributes embedded in this image.}
    \label{figure1}
\end{figure*}

However, a common challenge arises for these approaches: LR facial images can correspond to multiple HR facial images, which as a result would lead to inaccurate outputs; in other words, it could significantly change the identity of the hallucinated faces. For example, when hallucinating a LR facial image of a young person, the results may be a HR image of an old person due to this mapping ambiguity of LR-to-HR image hallucination. By incorporating facial attributes into the learning process and progressively resolving the facial image, the mapping between LR and HR images can be constrained more, reducing the ambiguity of the identity. Following this idea, we develop a face hallucination framework for upsampling. 
The proposed approach aims at utilizing a progressively trained upsampling and discriminative networks that predicts the facial attributes of a given image to super-resolve LR faces at all stages. The proposed framework includes an autoencoder, where the attribute vector is embedded at its bottleneck latent representation of key facial features. The autoencoder is followed by a series of deconvolutional layers and residual blocks to yield upsampled face outputs. These upsampled outputs are fed into several discriminative networks that determine the embedded attributes in the generated HR image and whether the generated HR image appears to be real or not. By doing so, the proposed model  super-resolves tiny (16$\times$16 pixels)  face images in a progressive manner until it is resolved by a factor of 8$\times$. Attribute regularization and adversarial progressive learning framework helps to reduce the ambiguity and uncertainty in the one-to-many mapping.

By progressively training the upsampling network\cite{karras2018progressive}, the proposed approach yields a novel architecture that progressively incorporates features from low to high in the process of upsampling the image. We conduct thorough evaluations with a large face dataset, CelebA \cite{liu2015faceattributes}, and that the proposed approach achieves satisfactory face hallucination results that outperforms other state-of-the-art methods.

The main contribution is summarized as follows:
\begin{itemize}
    \item The proposed approach is one of the first facial super resolving networks which utilize progressive training in conjunction with attributes for regularization and can also manipulate the hallucinated faces based on the facial attributes.
    \item The proposed approach enables a network that is not reliant on providing attribute information of the given LR input post training. 
\end{itemize}

The following sections of the paper are organized as follows. In Section \ref{related}, we review recent relevant works for face hallucination and super-resolution. In Section \ref{method}, we describe the details of the proposed approach. In Section \ref{exp}, we present both qualitative and quantitative results of the proposed approach along with  detailed ablation studies. Finally, the work is concluded in Section \ref{conclusion}.

\section{Related Work}\label{related}
Due to a large amount of related works in the literature, we briefly review a few recent relevant works in this following.\\

\noindent\textbf{Face Image Generation:} As face hallucination is a sub-field of image generation, many of the approaches between the two areas are similar. Goodfellow \emph{et al.}\cite{NIPS2014_5423} introduces generative adversarial networks (GAN) that are able to generate images from  noise vectors.
Many alternative GAN\cite{Arjovsky2017WassersteinG,NIPS2015_5773,Zhao2016EnergybasedGA} designs have also been introduced to improve the quality and resolution of generated images. 
Alongside generating images from noise vector, some approaches generate images from text-based inputs \cite{Zhang2016StackGANTT,Reed2016GenerativeAT}.\\

\noindent\textbf{Face Super-resolution:} According to Wang \emph{et al.}\cite{DBLP:journals/corr/abs-1902-06068}, in recent years, image super resolution has witnessed tremendous progress utilizing deep learning models. These deep learning models use a variety of approaches to super resolve faces. Models such as VDSR \cite{DBLP:journals/corr/KimLL15b} upsample the LR image before processing the image through a network. Other methods such as Yu \emph{et al.} \cite{Yu_2018_CVPR} utilize progressive upsampling, which refers to the image being upsampled throughout the network rather than being done in one shot at the beginning or the end.\\

\noindent\textbf{Facial Attribute Manipulation:} In addition to super resolution, the proposed approach also allows to perform attribute manipulation on the hallucinated images. Yu \emph{et al.} achieves this through feeding in a vector where each element represents an attribute and its presence in the given input into the bottleneck of the auto encoder, as well as feeds in this attribute vector into the discriminator. AttGAN \cite{8718508}, also feeds in an attribute vector into the bottleneck of the autoencoder within the Generative network, but its discriminative network tries to predict the attributes present in the input image rather than encoding the attributes in it. Other approaches that manipulate facial attributes include Shen \emph{et al.}, \cite{Shen2016LearningRI} which changes attributes of an input image through its residual image and Perarnau \emph{et al.},\cite{Perarnau2016InvertibleCG} which edits attributes of a given face through an invertible conditional GAN.  Lee \emph{et al.},  \cite{Lee2018AttributeAC} use attributes to stabilize local regions to help facial hallucination.\\

In more recent approaches, the distilled-FAN \cite{Kim2019ProgressiveFS} and Ahn \emph{et al.}, \cite{Ahn2018ImageSV} also utilizes a progressively trained network. Distilled-FAN utilizes facial landmarks to help constrain the mapping between LR and HR facial images, while Ahn \emph{et al.}, utilizes progressive learning and cascading residual networks to help with general image super resolution. Unlike these approaches based on shape and pixel information which are more sensitive to resolution, the proposed approach progressively infuses the attributes to guide the training process and more effectively mitigates the identity change issue during the process of LR-to-HR hallucination.

\section{The Proposed Approach}\label{method}

To handle these problems, we present an end-to-end upsampling network that utilizes LR images of faces and their corresponding facial attributes as inputs and produces a HR version of these faces as outputs. Due to the nature of progressively training, the network can reduce the ambiguity of the identity of the LR faces at each upsampled level (2$\times$,4$\times$,8$\times$), where the upsampling network takes the LR faces as inputs and embeds facial attributes, while constraining the mapping to HR face outputs at each scale. Each multi-task discriminative network simultaneously tries to predict the embedded attributes of each HR face, both generated and ground truth face images as well as determine whether an image is a ground truth image or if the image is generated by the upsampling network.
The overview of the system pipeline is shown in Figure\ref{figure1}. In the following subsections, we describe each component in detail.

\subsection{Upsampling Network}

The upsampling network is comprised of the attribute embedding  convolutional autoencoder and up-scaling layers with residual blocks in between the final layers. The attributes are first encoded as $n_a$-dimensional  vector with each value ranging between 0 and 1 where $n_a$ is the number of attributes and concatenated with the latent representation of the bottleneck of the autoencoder. The inclusion of attribute embedding allows the incorporation of the semantic information for faces to constrain the subject identity while mapping from the LR image space to the HR one. For the encoding part of the network, leaky Rectifier Linear Units (lReLU) are  utilized, while in the upsampling and decoding portion the ReLU activation function is used. In between each layer, batch normalization is applied to the outputs of the previous layer. 

In addition, the input of each residual block is fed into both residual block and a RGB block. The RGB block takes this as input and outputs an image with the same dimension of output for the current stage. Then, this output is recursively added to an upsampled version from the RGB block of the previous layer. At the first stage (which produces a 32$\times$32 image), the output of the RGB block is added to the upsampled version of the LR input of the network. After the first up-sampling layer, it hallucinates the input to a 32$\times$32 pixel version of the target image, and so on until it reaches the resolution of 128$\times$128 pixels. Residual blocks are included in our model as they not only help the optimization of the network but also allow for a deeper network \cite{DBLP:journals/corr/HeZRS15}. This approach works well when progressively training the network since at each stage the network is trying to map the LR input to the corresponding down-sampled version of the HR image.

On the other hand, since the network is progressively trained, the highest resolved resolution is increased after a certain amount of epochs. To constrain these outputs to be consistent, the network utilizes pixel-wise loss in the $\ell_{1}$ norm to a bilinearly downsampled version of the ground truth HR image. For example, if the network is currently resolving the image to 32$\times$32 pixels, or 2$\times$ the LR input, then the image that would be used to constrain the output would be a 32$\times$32-pixel bilinearly downsampled version of the ground truth HR image. This is done all the way to 128$\times$128 pixels, 8$\times$ the LR input, where in addition to pixel-wise loss the network also employs perceptual loss \cite{Johnson2016PerceptualLF}. Perceptual loss measures the Euclidean distance between the the image's features and helps constrain the face semantics of the upsampled HR image and the ground truth HR image to be close. We exploit the VGG-Face \cite{Parkhi15}, based on \cite{Simonyan2014VeryDC}, model to extract the image features related to subject identity. 

Besides these two losses, the upsampling network also includes a binary cross entropy loss between predicted attributes utilized to create a given image and those attributes fed into the upsampling network. Finally, there is an adversarial loss function, and we employed the one used in Wassterstein Generative Adversarial Network (WGAN)\cite{Arjovsky2017WassersteinG}, where for the generative network the adversarial loss is the negative mean of the adversarial prediction of the discriminator. 

\subsection{Discriminative Network}

Similar to \cite{DBLP:journals/corr/YanYSL15}, the goal of the discriminative networks is to
ensure the upsampling network is faithfully augmenting the LR facial image and incorporates facial semantic features. The proposed approach does this by utilizing a discriminative network with branched outputs. In particular, the discriminative network is utilized to determine whether the facial attributes of the upsampled images are faithful to the encoded attributes provided to the upsampling network as well as constrain these upsampled images to their HR ground truth faces. 

While the upsampling network encodes these attributes, within the bottleneck of the autoencoder, the network could potentially find them meaningless or ignore them, this would mean their weights could be zero. To avoid this, the discriminative network needs to be able to reinforce the attributes during the generative process. 
Another aspect to note is that the model uses a progressive learning approach. In the progressive learning approach the output of the upsampling network is not consistent, and grows over the course of the training process. To address the upsampling model's progressively growing output, our discriminative network needs to either incorporate an additional input layer at ever stage or employ multiple discriminative networks.

The model utilizes multiple discriminative networks, one for each size output for the upsampling network, e.g. one that takes in a 32$\times$32 pixel image, one for 64$\times$64 pixel images, and one for 128$\times$128 pixel images. 
The reason our model employs multiple discriminative networks rather than one discriminative network that is progressively grown is that during the training process the discriminative network is updating at a faster rate than the upsampling network. This is done to allow the discriminative network to learn to predict the present attributes in the image. Since the discriminative network is updated at a faster rate than the upsampling network, a progressively grown discriminative network would over power the upsampling network. This is expected as the discriminative network would more efficiently determine the real images from fake images sooner, and as such would stunt the growth of the upsampling network.

\subsection{Attribute Classifier}

The model also includes an attribute classifier, which takes in an LR facial image and tries to predict the attributes that are present within the given image.This network addresses the issue of unknown attributes for unseen LR face images as the attribute classifier will be able to learn the attributes of these new faces.This network also outputs an $n_a$-dimensional vector with each value ranging between 0 and 1 for each attribute, essentially representing the presence a given attribute within an image. Using this network to classify attributes of new LR images in the test set, our model is able to perform relatively the same as if our upsampling network is being fed in the ground truth attributes. Since this network is a binary classifier, it is trained with binary cross entropy loss. 

\subsection{Training}
This facial super resolution network is trained using a progressive learning approach. The upsampling network is denoted as $G_i$, where i denotes what stage it is on, e.g. $G_{1}$ represents the 32$\times$32 pixel intermediary output of the network, $G_{2}$ for 64$\times$64, etc.  The discriminative networks have branched outputs, so the prediction for the embedded attribute will be denoted as $D^{Attr}_{i}$ while the adversarial output is denoted as $D^{Adv}_{i}$. Since there are multiple discriminative networks i represents what staged each network is used at. The LR face images are denoted as $I^{LR}$ and the binary ground truth attribute vectors of these images are denoted as $a$. The target HR face images are denoted as $I^{HR}_{i}$, where \emph{i} represents the stage at which this HR image is the target. The HR images reconstructed by the upsampling network are denoted as $G_{i}(I^{LR}, a)$. The outputs of the upsampling network where randomized attributes, denoted as $a^*$, are given as inputs along with $I^{LR}$ are denoted as $G_{i}(I^{LR}, a^*)$. Since the goal of this network is to super resolve LR to HR images, the ground truth attributes are only used as labels for their corresponding image.

The upsampling network is trained using perceptual loss, a pixel-wise $\ell_{1}$ loss, an adversarial loss from the discriminator networks, and finally an attribute loss from the discriminative networks. One thing to note about the adversarial and attribute loss is that these losses are calculated based on LR facial inputs that are given randomized attributes. This ensures that the upsampling network would be able to generate these attributes without linking certain attributes together due to reoccurring patterns with these attributes, i.e. the upsampling network may only generate makeup or sunglasses if the the attribute of being female is present. In addition, this approach allows for the upsampling network to learn different combinations of attributes not present within the dataset. The pixel wise and perceptual losses are both calculated using the upsampling networks reconstructed from the ground truth, as these outputs have ground truth images to compare with, which is needed for evaluating pixel wise and perceptual losses. The loss function we are trying to minimize at each stage is as follows: 

\begin{equation}\label{eq1}
\begin{split}
L_{G_i} = -\mathbb{E}[\mathbf{G}_{i}(I^{LR}, a^*)] + \alpha || \mathbf{G}_{i}(I^{LR}, a) - I^{HR}_i ||\\ - \beta (\mathbb{E}[log(\mathbf{D}^{Attr}_{i}(\mathbf{G}_{i}(I^{LR}, a^*))]\\ + \mathbb{E}[1-log(\mathbf{D}^{Attr}_{i}(\mathbf{G}_{i}(I^{LR}, a^*))])\\+ \gamma || \theta(\mathbf{G}_{i}(I^{LR}, a)) - \mathbf{\theta}(I^{HR}_i)||
\end{split}
\end{equation}

It should be noted that $\theta(\cdotp)$ represents the feature vectors extracted from layer "ReLU41" from VGG-Face.

The discriminators are updated more frequently than the upsampling network, as the discriminators are trying to learn the attributes of the images and the upsampling network relies on this loss, as well as the adversarial loss, to update its own weight.

For each 2$\times$ super resolution, there is a corresponding discriminator network and a corresponding output from the upsampling network. Thus, there are three discriminator one for 2$\times$, one for 4$\times$, and one for 8$\times$.
The discriminator has two outputs, one a one-dimensional vector that predicts the present facial attributes within the given image, and second an adversarial loss that predicts likely the given image is real or fake. The discriminator is updated using the loss present in WGAN, which is the negative difference in the means of the prediction of the generated image and the ground truth image. In addition to this, there is a gradient penalty utilized to keep the Wasserstein distance close to 1 which helps to improve the training of WGANs \cite{DBLP:journals/corr/GulrajaniAADC17,wei2018improving}. Finally, the last loss function incorporated is the binary cross entropy loss between the predicted attributes and the ground truth attributes. The loss function that each discriminator is trying to minimize is as follows:
\begin{equation}\label{eq2}
\begin{split}
L_{D_i} = -(\mathbb{E}[\mathbf{D}^{Adv}_{i}(I^{HR}_i)] - \mathbb{E}[\mathbf{D}^{Adv}_{i}(\mathbf{G}_{i}(I^{LR}, a))])\\ + (\mathbb{E}[log(\mathbf{D}^{Attr}_{i}(\mathbf{G}_{i}(I^{LR}, a))]\\ + \mathbb{E}[1-log(\mathbf{D}^{Attr}_{i}(\mathbf{G}_{i}(I^{LR}, a))])\\ + \lambda \mathbb{E}[(||\bigtriangledown_{\hat{I}^{HR}_i}\mathbf{D}(\hat{I}^{HR}_i)||_2 -1)^2]
\end{split}
\end{equation}
It should be noted in \eqref{eq2}, $\hat{I}^{HR}_i$ represents:

\begin{equation}
\hat{I}^{HR}_i = t\times I^{HR}_i + (1-t)(\mathbf{G}_{i}(I^{LR}, a)),  0 \leq t \leq 1
\end{equation}

\noindent where $\hat{I}^{HR}_i$ is only used in the WGAN gradient penalty for the discriminators.

The attribute classifier is updated using the binary cross entropy loss between its prediction and the ground truth attributes for the LR face image passed through the classifier. The loss function for the attribute classifier is as follows:
\begin{equation}\label{eq3}
\begin{split}
L_A = -\sum_{j=1}^{m}\sum_{i=1}^{n}a_{i}log(\mathbf{A}_{i}(I_{j}^{LR})) \\+(1-a_i)log(1-\mathbf{A}_{i}(I_{j}^{LR}))
\end{split}
\end{equation}

Where a is the attribute and calculates the binary cross entropy loss for each attribute, denoted by i, and ${A}_{i}(I^{LR})$ is the predicted probability of the ith attribute. This loss is calculated over all j training samples, and their  i attributes. 

For the convolutional layers in the Generator, Discriminator, and Attribute Classifier, we use a filter size of 4$\times$4, a stride size of 2 and a padding of 1. The only exception to this are the convolutional layer used to produce the intermediary output which use a filter size of 5$\times$5, a stride size of 1 and a padding size of 1. In addition, there are hyper parameters for each of the losses in the generative model. In our experiments, $\alpha$ is set to 100, $\beta$ is set to 10 and $\gamma$ is set to .1. The hyper parameter $\lambda$ used in the discriminator is set to 10. The attributes that this network is trying to predict the presences of is as follows: Bald, Bangs, Black Hair, Blond Hair, Brown Hair, Bushy Eyebrows, Eyeglasses, Male, Mouth Open, Mustache, Pale, Young. For a more detailed network architecture, please refer to Appendix \emph{1}, in supplemental material.
\begin{figure*}[t]
    \centering
    \begin{subfigure}{0.065\textwidth}
    \includegraphics[width=\textwidth]{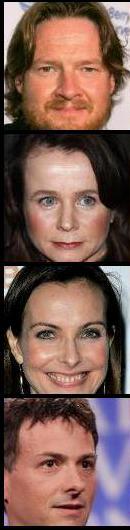}
    \caption{}
    \label{fig:gt}
    \end{subfigure}
    \begin{subfigure}{0.065\textwidth}
    \includegraphics[width=\textwidth]{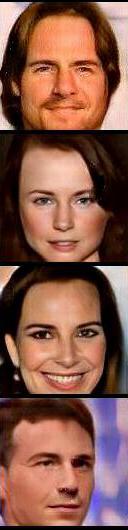}
    \caption{}
    \label{fig:g}
    \end{subfigure}
    \begin{subfigure}{0.065\textwidth}
    \includegraphics[width=\textwidth]{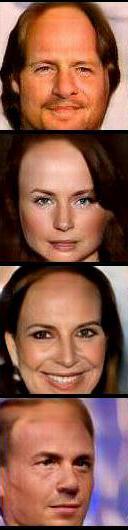}
    \caption{}
    \label{fig:bald}
    \end{subfigure}
    \begin{subfigure}{0.065\textwidth}
    \includegraphics[width=\textwidth]{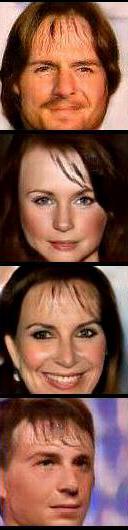}
    \caption{}
    \label{fig:bang}
    \end{subfigure}
    \begin{subfigure}{0.065\textwidth}
    \includegraphics[width=\textwidth]{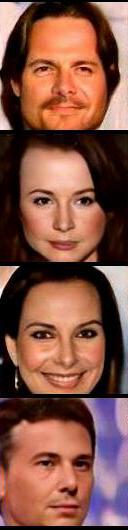}
    \caption{}
    \label{fig:black}
    \end{subfigure}
    \begin{subfigure}{0.065\textwidth}
    \includegraphics[width=\textwidth]{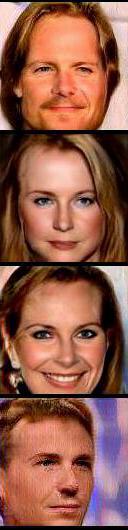}
    \caption{}
    \label{fig:blond}
    \end{subfigure}
    \begin{subfigure}{0.065\textwidth}
    \includegraphics[width=\textwidth]{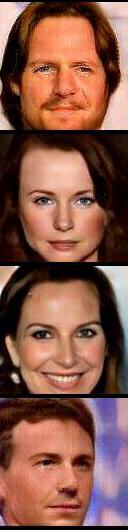}
    \caption{}
    \label{fig:brown}
    \end{subfigure}
    \begin{subfigure}{0.065\textwidth}
    \includegraphics[width=\textwidth]{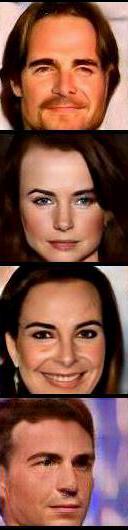}
    \caption{}
    \label{fig:bushy}
    \end{subfigure}
    \begin{subfigure}{0.065\textwidth}
    \includegraphics[width=\textwidth]{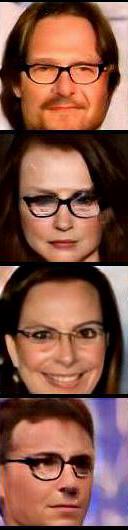}
    \caption{}
    \label{fig:glasses}
    \end{subfigure}
    \begin{subfigure}{0.065\textwidth}
    \includegraphics[width=\textwidth]{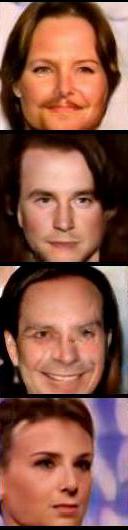}
    \caption{}
    \label{fig:male}
    \end{subfigure}
    \begin{subfigure}{0.065\textwidth}
    \includegraphics[width=\textwidth]{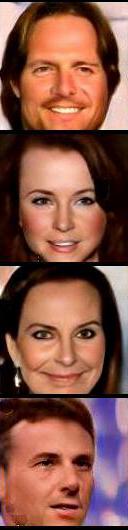}
    \caption{}
    \label{fig:mouth}
    \end{subfigure}
    \begin{subfigure}{0.065\textwidth}
    \includegraphics[width=\textwidth]{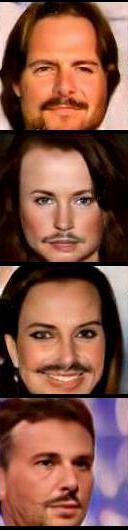}
    \caption{}
    \label{fig:mustache}
    \end{subfigure}
    \begin{subfigure}{0.065\textwidth}
    \includegraphics[width=\textwidth]{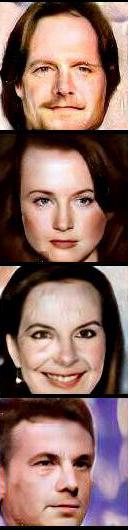}
    \caption{}
    \label{fig:pale}
    \end{subfigure}
    \begin{subfigure}{0.065\textwidth}
    \includegraphics[width=\textwidth]{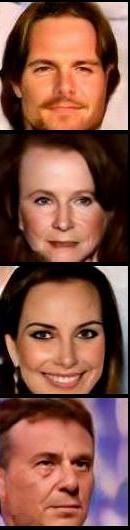}
    \caption{}
    \label{fig:young}
    \end{subfigure}
    \caption{This figure depicts our network's capabilities of manipulating attributes of the reconstructed face. Each column represents an attribute of that is altered to be different from the original attributes of that face. (a) Original Image. (b) Upsampled Image produced by the network. (c) Bald. (d) Bangs. (e) Black Hair. (f) Blond Hair. (g) Brown Hair. (h) Bushy Eyebrows. (i) Eyeglasses. (j) Swaps the sex. (k) Mouth Open. (l) Mustache. (m) Pale. (n) Young.  }
    \label{fig:manip}
\end{figure*}
\section{Experimental Results}\label{exp}
We evaluate and compare our network with other state-of-the-art methods using qualitative and quantitative methods  \cite{Yu_2018_CVPR,DBLP:journals/corr/KimLL15b,Wang_2018_CVPR_Workshops}, and we also conduct ablation studies to investigate the usefulness of each component of the proposed approach. Yu \emph{et al.} \cite{Yu_2018_CVPR}, utilize attributes provided by the CelebA dataset and encodes these attributes into the bottleneck of the autoencoder. In addition, their network also embeds attributes into the discriminative network to help enforce attribute learning. Kim \emph{et al.} \cite{DBLP:journals/corr/KimLL15b} present a super resolution method that uses a very deep CNN approach. Wang \emph{et al.} \cite{Wang_2018_CVPR_Workshops} discuss another super resolution approach that also takes a progressive approach to super-resolve images  and introduces novel super resolution network and a super resolution GAN. 
\begin{figure*}[t]
    \centering
    \begin{subfigure}{0.1\textwidth}
    \includegraphics[width=\textwidth]{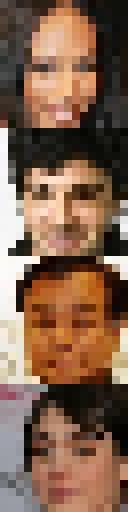}
    \caption{}
    \label{fig:con_nn}
    \end{subfigure}
    \begin{subfigure}{0.1\textwidth}
    \includegraphics[width=\textwidth]{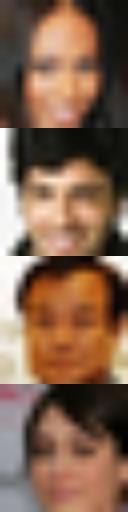}
    \caption{}
    \label{fig:con_b}
    \end{subfigure}
    \begin{subfigure}{0.1\textwidth}
    \includegraphics[width=\textwidth]{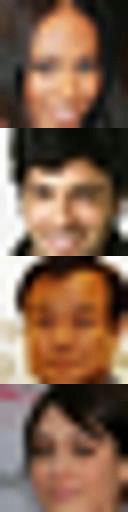}
    \caption{}
    \label{fig:con_vdsr}
    \end{subfigure}
    \begin{subfigure}{0.1\textwidth}
    \includegraphics[width=\textwidth]{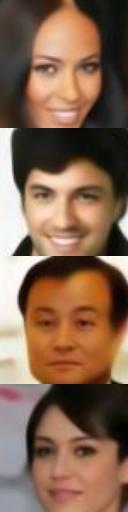}
    \caption{}
    \label{fig:con_proSR}
    \end{subfigure}
    \begin{subfigure}{0.1\textwidth}
    \includegraphics[width=\textwidth]{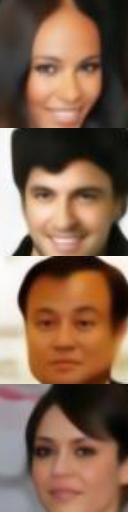}
    \caption{}
    \label{fig:con_proSRGAN}
    \end{subfigure}
    \begin{subfigure}{0.1\textwidth}
    \includegraphics[width=\textwidth]{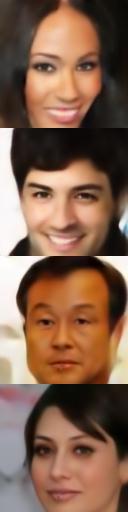}
    \caption{}
    \label{fig:con_yu}
    \end{subfigure}
    \begin{subfigure}{0.1\textwidth}
    \includegraphics[width=\textwidth]{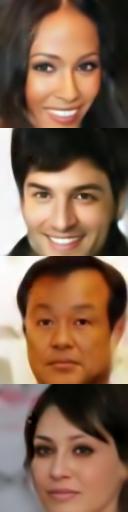}
    \caption{}
    \label{fig:con_attr}
    \end{subfigure}
    \begin{subfigure}{0.1\textwidth}
    \includegraphics[width=\textwidth]{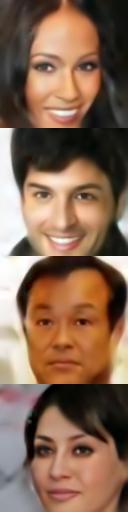}
    \caption{}
    \label{fig:con_ours}
    \end{subfigure}
    \begin{subfigure}{0.1\textwidth}
    \includegraphics[width=\textwidth]{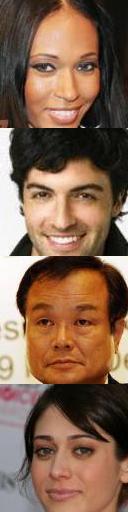}
    \caption{}
    \label{fig:con_gt}
    \end{subfigure}
    \caption{Qualitative comparison between our results and other state of the art methods. (a) LR input (b) Bilinear. (c) VDSR. (d) ProSR. (e) ProGanSR. (f) Yu's. (g) Ours with the attributes from the attribute classifier. (h) Ours with ground truth attributes.  (i) Target.}
    \label{fig:concat}
\end{figure*}

\subsection{Dataset}
The dataset used to train this network is the Large Scale CelebFaces Attributes (CelebA) dataset \cite{liu2015faceattributes}. Then images from the dataset are then centered-cropped to 120$\times$120 pixels and resized to 128$\times$128 pixel, which will be used as the HR images. These HR images are then down-sampled until they are 16$\times$16 pixel to represent the LR images. The training set is comprised of 182k images from the dataset and their corresponding HR, LR images, and ground truth attributes. There are a select few attributes used in the training process and are extracted for each image. These extracted attributes are treated as the ground truth attributes. The testing set is comprised of the remaining 20K images of CelebA. Our model is trained on the aligned version of the CelebA, all the images of the faces are upright and aligned. 

\begin{table*}[h!]
\caption{Quantitative comparison with SoA methods over test dataset}
\begin{center}

\begin{tabular}{|c|c|c|c|c|c|c|c|}
\hline
Method & Bilinear & VDSR & ProSR & ProGanSR & Yu  & Ours with Attr. Class& Ours with G.T. Attr.\\
\hline
PSNR & 20.75& 21.65  & 22.37 & 22.56& 21.82 & 22.76 &\textbf{22.89} \\
\hline
SSIM & 0.574 & 0.652 & 0.607  &0.614 &0.659      & 0.6804 &\textbf{0.6814}\\
\hline
\end{tabular}
\end{center}
\label{tab:comp}
\end{table*}

\subsection{Comparison with Other State-of-the-Art}

We use the average PSNR score and the structural similarity score (SSIM) to compare the the performance of our method other state-of-the art methods. Table\ref{tab:comp} shows that our method performs better than these methods, even when utilizing the attribute classifier to feed into our network. Yu \emph{et al.},\cite{Yu_2018_CVPR} also utilize attributes but do not progressively train their model, indicating that progressive training of a model does help with face hallucination tasks. When comparing to more recent models, such as distilled FAN's \cite{Kim2019ProgressiveFS}, our model achieves a higher PSNR score and a similar SSIM score. Their scores are 22.66 and 0.685 for PSNR and SSIM respectively. This is only for their model's performance on aligned images from CelebA. The incorporating of facial landmarks in conjunction with attributes could yield better results in future works.

When we compare our results qualitatively as shown in Figure \ref{fig:concat}, we see that the proposed method is able to faithfully restore certain attributes of images that are generally lost in the face hallucination process. An example is in Table \ref{tab:comp} when comparing the older male in the third row. Upon checking at the results, we find that there is some restoration of wrinkles and aged looks. When comparing to the results from ProSR and ProSRGAN \cite{Wang_2018_CVPR_Workshops}, we see that the faces are overly smoothed and as such lose this detailed depiction of the ground truth image when upsampling the image. In the case of VDSR \cite{DBLP:journals/corr/KimLL15b}, the images are very blurry, this can be accounted for the fact that VDSR is trained on natural images, and as such can not fully capture facial structure and details. From qualitative observations at these results, we find that the inclusion of attributes helps in the restoration of facial attributes. Finally, when comparing our results with Yu \emph{et al.}, we see that our approach has less blurriness/white tint in the final results  suggesting progressive training is effective for the face hallucination process. 

\begin{figure}[t]
    \centering
    \includegraphics[height=5cm]{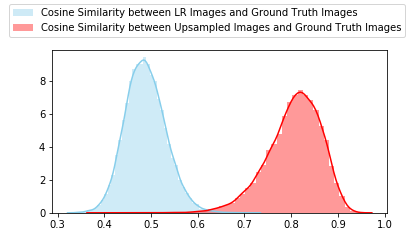}
    \caption{This figure compares the cosine similarities between the feature vectors between LR images and ground truth versus upsampled images and the ground truth. These feature vectors are extracted from VGG-Face\cite{Parkhi15} layer "ReLU41". Only the test dataset is used in this comparison. }
    \label{fig:cosine}
\end{figure}

\subsection{Attribute Manipulation}

To the best of our knowledge, most previous methods that super resolve LR facial images do not have the ability to edit certain attributes of the LR images \cite{Kim2019ProgressiveFS,Wang_2018_CVPR_Workshops,DBLP:journals/corr/ZhuLLT16}. Similar to \cite{Yu_2018_CVPR}, our method utilizes attributes to upsample LR images to HR images. Hence, our methods can also edit the attributes of the generated HR face images, such that the attributes of HR facial image can differ from those of the LR facial image used to resolve them. As shown in Figure \ref{fig:manip}, our model is able to manipulate a variety of attributes such as make-up, how pale someone is, as well as the color of their hair. 

\begin{figure*}[t]
    \centering
    \begin{subfigure}{0.1\textwidth}
    \includegraphics[width=1\textwidth,height=6.5cm]{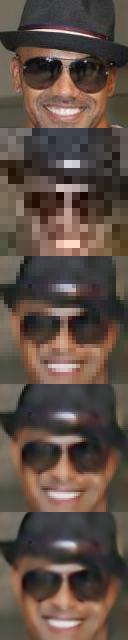}
    \caption{}
    \label{fig:img1}
    \end{subfigure}
    \begin{subfigure}{0.12\textwidth}
    \includegraphics[width=\textwidth,height=6.5cm]{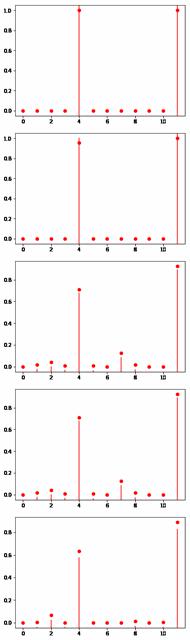}
    \caption{}
    \label{fig:attr1}
    \end{subfigure}
    \begin{subfigure}{0.1\textwidth}
    \includegraphics[width=\textwidth,height=6.5cm]{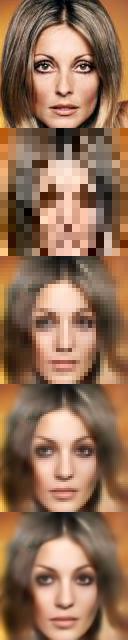}
    \caption{}
    \label{fig:img2}
    \end{subfigure}
    \begin{subfigure}{0.12\textwidth}
    \includegraphics[width=\textwidth,height=6.5cm]{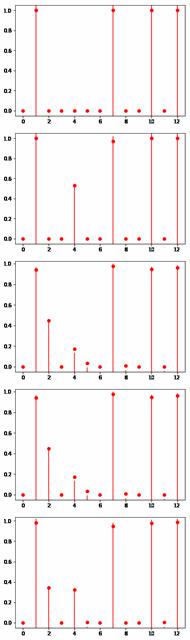}
    \caption{}
    \label{fig:attr2}
    \end{subfigure}
    \begin{subfigure}{0.1\textwidth}
    \includegraphics[width=\textwidth,height=6.5cm]{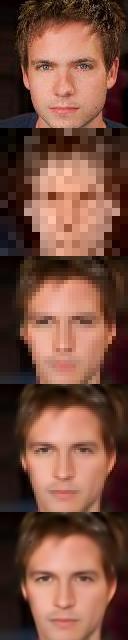}
    \caption{}
    \label{fig:img3}
    \end{subfigure}
    \begin{subfigure}{0.12\textwidth}
    \includegraphics[width=\textwidth,height=6.5cm]{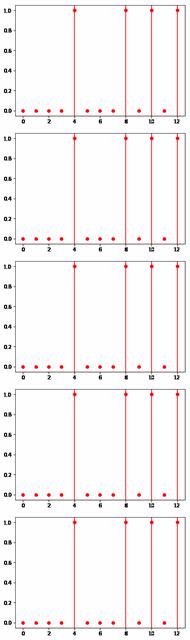}
    \caption{}
    \label{fig:attr3}
    \end{subfigure}
    \begin{subfigure}{0.1\textwidth}
    \includegraphics[width=\textwidth,height=6.5cm]{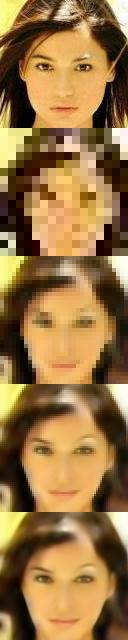}
    \caption{}
    \label{fig:img4}
    \end{subfigure}
    \begin{subfigure}{0.12\textwidth}
    \includegraphics[width=\textwidth,height=6.5cm]{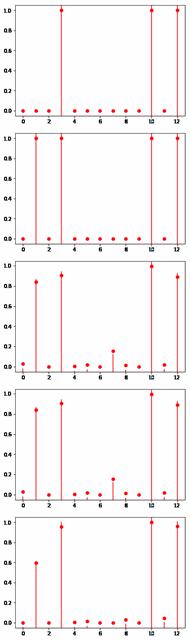}
    \caption{}
    \label{fig:attr4}
    \end{subfigure}
    \caption{  This figure shows how consistent the discriminators are in predicting the present attributes across all stages. Columns with face images are organized from top to bottom in the following manner: Ground Truth image, LR input, 32$\times$32 pixel intermediary output, 64$\times$64 pixel intermediary output, 128$\times$128 pixel output. Columns to the right of columns with faces are that stage's discriminator's prediction of the present attributes, the only except to this is the top most graph which is the ground truth attributes, present in the ground truth image. }
    \label{fig:attr}
\end{figure*}

\begin{figure*}[h!]
    \centering
    \begin{subfigure}{0.1\textwidth}
    \includegraphics[width=\textwidth]{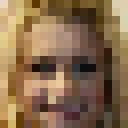}
    \caption{}
    \label{fig:ab_nn}
    \end{subfigure}
    \begin{subfigure}{0.1\textwidth}
    \includegraphics[width=\textwidth]{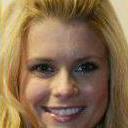}
    \caption{}
    \label{fig:con_g}
    \end{subfigure}
    \begin{subfigure}{0.1\textwidth}
    \includegraphics[width=\textwidth]{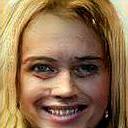}
    \caption{}
    \label{fig:ab_or}
    \end{subfigure}
    \begin{subfigure}{0.1\textwidth}
    \includegraphics[width=\textwidth]{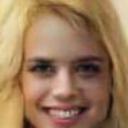}
    \caption{}
    \label{fig:ab_no_pro_per}
    \end{subfigure}
    \begin{subfigure}{0.1\textwidth}
    \includegraphics[width=\textwidth]{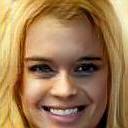}
    \caption{}
    \label{fig:ab_no_pro}
    \end{subfigure}
    \begin{subfigure}{0.1\textwidth}
    \includegraphics[width=\textwidth]{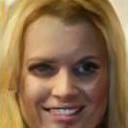}
    \caption{}
    \label{fig:ab_no_attr}
    \end{subfigure}
    \begin{subfigure}{0.1\textwidth}
    \includegraphics[width=\textwidth]{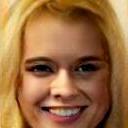}
    \caption{}
    \label{fig:ab_attr}
    \end{subfigure}
    \caption{Ablation study of the proposed approach (a) LR input (b) HR target. (c) Results when only using adversarial loss. (d) Results when pixel wise loss is introduced . (e) Results when perceptual loss is introduced. (f) Results when network is progressively trained. (g) Results when attributes are introduced.}
    \label{fig:ablation}
\end{figure*}

\subsection{Discussion}

First, we conduct the ablation study to thoroughly investigate the effectiveness of each component by incrementally introducing them into the framework. As shown in Figure~\ref{fig:ablation}, when we use all the proposed losses, the result of Figure \ref{fig:ablation}g is the closest one to the ground truth Figure \ref{fig:ablation}b. Next, for more qualitative analysis, we want to assess whether hallucinated HR images more accurately portray the features with ground truth images compared to their LR ones. To do this, we compare the cosine similarities between the feature vectors of the LR image and ground truth versus the upsampled HR image and ground truth where the feature vectors are extracted from the layer "ReLU41" of the off-the-shelf VGG-FACE model \cite{Parkhi15}. As shown in Figure \ref{fig:cosine}, we notice that the cosine similarities between the upsampled HR images and ground truth is much closer to 1 than their LR counterparts. This indicates that the proposed method does effectively address the challenge of LR images not fully encapsulating the features of their HR counterpart.

In addition to this, we also look at how consistent our discriminator's prediction of attributes across all stages. Figure \ref{fig:attr} shows examples of test images and how well our network does in predicting attributes. Columns \ref{fig:img1}, \ref{fig:img2}, \ref{fig:img3}, and \ref{fig:img4} are images of faces at varying scales, starting with the LR input all the way to the final output. The exception to this is that the top image is the ground truth image. To the right of each face column is its corresponding attribute prediction at each stage of the network. It should be noted that the order in which attribute is the same as how these attribute are manipulated in Figure \ref{fig:manip}, so the zeroth index is bald, first is bangs, etc. In some cases, such as columns \ref{fig:img3} and \ref{fig:attr3},  the discriminators do a great job of predicting the values of attributes to that of the ground truth. Interestingly, there is disagreement between the predictions of the discriminators and ground truth labels. Examples of this are in columns \ref{fig:img2} and \ref{fig:attr2}.
At the top of column \ref{fig:attr2} the ground truth attribute histogram indicates no presence of brown hair in its corresponding image in column \ref{fig:img2}. However, our model's discriminators across most stages predict there is some degree of brown hair present in the images in column \ref{fig:img2}. From looking at those images, it is definitely apparent that this is a matter of opinion as the hair does seem to have some blond highlights at the bottom but darkish brown at other places. Overall, our discriminators seem to be consistent in the attributes they predict to be present within each stage.

\section{Conclusion}\label{conclusion}

In this paper, we introduces an end-to-end LR-to-HR face hallucination network which  super-resolves LR images (16$\times$16 pixels) to the corresponding HR images (128$\times$128 pixels) using a progressive learning approach. Using a multi-task discriminator for attribute prediction and real/fake classification, the generative network yields higher quality HR facial images with attributes consistent with LR images. Furthermore, this network also allows for manipulation of the attributes.

{\small
\bibliographystyle{ieee}
\bibliography{references}
}
\end{document}